# Approximate Separability for Weak Interaction in Dynamic Systems

**Avi Pfeffer**
Harvard University

## Abstract

One approach to monitoring a dynamic system relies on decomposition of the system into weakly interacting subsystems. An earlier paper introduced a notion of weak interaction called separability, and showed that it leads to exact propagation of marginals for prediction. This paper addresses two questions left open by the earlier paper: can we define a notion of approximate separability that occurs naturally in practice, and do separability and approximate separability lead to accurate monitoring? The answer to both questions is affirmative. The paper also analyzes the structure of approximately separable decompositions, and provides some explanation as to why these models perform well.

## 1 Introduction

Monitoring the state of a dynamic system in the face of uncertainty is a central AI task. To perform this task, we need compact probabilistic representations of dynamic systems, and efficient and accurate monitoring algorithms. Dynamic Bayesian networks (DBNs) [Dean and Kanazawa, 1989] provide a compact representation by exploiting conditional independencies that hold between variables. However, exact monitoring in DBNs is hard, and approximate algorithms are needed.

One approach, introduced by Boyen and Koller (BK) [1998], is to factor the variables of the DBN into subsets. The joint distribution over all the variables is then approximated by the product of marginal distributions over the factors. This approach relies on the intuition that a complex system can be decomposed into weakly interacting subsystems. But what exactly does weak interaction mean? A standard way to frame this question is, when is the true distribution approximately the product of the factored distributions?

One idea is that two systems are weakly interacting if they are conditionally independent of each other given a small set of interface variables. This works well for static models. But for dynamic models, short-run independence does not lead to long-run independence and factors that satisfy this criterion may become highly dependent. Other ideas are suggested by BK, who provide bounds for the error obtained using their approximation in terms of several quantities. A crucial one is the mixing rate of the system, which intuitively is the rate at which the system "forgets" old errors. While this is important for determining when the method will work well, it is a property of the system as a whole, not a characterization of weak interaction. Two other quantities they introduce are the number of other factors a factor depends on, and more importantly the number it influences. In [Boyen and Koller, 1999], they extend their analysis and introduce a measure of the degree to which factors forget the influence of other factors. However the bounds they obtain using these quantities are quite weak and not reflective of how the method performs in practice.

In an earlier paper [Pfeffer, 2001], we presented a new notion of weak interaction based on a different question. The new question is, when are the factor distributions obtained using the BK approach correct? This may occur even though the joint distribution is far from being a product of the factor distributions. We defined a criterion called separability, saying that the conditional probability distribution for a factor is separable if it can be decomposed additively into terms depending on each of the factors individually. We showed that if the factors are separable, marginal distributions can correctly be propagated from one time step to the next. However, observations break this property, so the BK approach with separable models only produces exactly correct marginals for prediction, not monitoring.

That paper left open two important questions. The first is whether separable models actually perform well at the monitoring task. The reason to believe that might be the case is that they correctly propagate marginal distributions through the dynamics, and only introduce error in conditioning on observations, unlike non-separable models which have both kinds of error. The second question asks whether approximating the separability property can lead to more natural models that still have good performance. This paper addresses both these questions.

In this paper we define approximate separability, and show that it is a natural property that can hold for real models. We show experimentally, on some simple examples, that separability and approximate separability indeed lead to good performance on both the prediction and monitoring tasks. Thus approximate separability, at least on these examples, appears to be a sufficient condition for making BK work well. We also show that separability can be used to produce a better factorization of a system than the "obvious" one. We analyze the structure of approximately separable models, presenting a method for automatically determining the degree of separability of a conditional probability distribution. Finally, we address the question of why separable and approximately separable models perform so well on the monitoring task. First we present a bound on the monitoring error of a separable model for a simple special case. We then define two different possible sources of error that may arise using the BK approach, and show that one of them is much less important than the other. As it turns out, separable models only have the less important kind of error in the simple examples studied, which goes some way towards explaining why they perform well in these examples.

## 2 Preliminaries

We will assume that the reader is familiar with dynamic Bayesian networks (DBNs) [Dean and Kanazawa, 1989; Murphy, 2002]. DBNs provide a compact representation of a dynamic model. However, monitoring in DBNs is difficult, because even though variables are conditionally independent in the short run they generally become dependent in the long run. As a result, in order to do exact monitoring one needs to maintain a joint distribution over all the state variables, which is infeasible. Therefore approximate monitoring algorithms are needed. One approach to approximate monitoring was introduced by Boyen and Koller [1998]. In their approach, the set of variables $\mathbf{X}$ is divided into factors $\mathbf{X}_1, ..., \mathbf{X}_m$, each of which consists of a subset of the variables.[1] Instead of maintaining a complete joint belief state $\mu(\mathbf{X})$, marginal distributions $\hat{\mu}_{\mathbf{X}_i}(\mathbf{X}_i)$ are maintained. The marginals are propagated from one time step to the next using a junction tree constructed from the two-time-slice model of the DBN, in which every factor $\mathbf{X}_i$ and $\mathbf{X}_i^-$ is contained in a clique. (The notation $\mathbf{X}_i^-$ means the variable $\mathbf{X}_i$ at the previous time point.) The joint distribution is approximated as $\hat{\mu}(\mathbf{X}) = \prod_{i=1}^m \hat{\mu}_{\mathbf{X}_i}(\mathbf{X}_i)$. BK provide a bound on the expected KL distance between $\mu$ and $\hat{\mu}$, and show that it does not grow over time.

In a previous paper [Pfeffer, 2001], we asked under what circumstances does the BK approach work well. Rather than focus on the difference between $\mu$ and $\hat{\mu}$, however, we asked when is the marginal $\hat{\mu}_{\mathbf{X}_i}$ equal to the true marginal $\mu_{\mathbf{X}_i}$. To answer this question, we introduced notions of sufficiency and separability.

**Definition 2.1:** Let $X$,$Y$ and $Z$ be three variables, and let $P(Z|XY)$ be given. $P(Z|XY)$ defines a mapping $\Phi : \Delta_{XY} \rightarrow \Delta_Z : \Phi(\pi) = \sum_{xy} \pi(xy)P(Z|xy)$. $X$ and $Y$ are *sufficient* for $Z$ if $\Phi$ depends only on the marginals of $\pi$ over $X$ and $Y$. ∎

In other words, sufficiency says that if we want to know $P(Z)$, we only need to know $\pi(X)$ and $\pi(Y)$ and do not need to know the joint $\pi(XY)$. Separability is defined as follows:

**Definition 2.2:** $P(Z|XY)$ is *separable* if there exist $\gamma$ and conditional probability distributions (CPDs) $P_X$ and $P_Y$, such that $P(Z|XY) = \gamma P_X(Z|X) + (1 - \gamma)P_Y(Z|Y)$. ∎

These definitions can naturally be generalized to more than two parents. It turns out that these two properties are equivalent: $X$ and $Y$ are sufficient for $Z$ if and only if $P(Z|XY)$ is separable. Note that there is no requirement that $\gamma$ be between 0 and 1. This is in contrast to the previous paper, which required that $\gamma$ be between 0 and 1. The statement that separability and sufficiency are equivalent definitely holds for the definition given here. The proof in [Pfeffer, 2001] implies that it also holds for the stricter definition. The two statements together imply that whenever a CPD has a separable decomposition with gamma not between 0 and 1, it also has one with gamma between 0 and 1. Unfortunately we have not been able to prove this directly, which leaves open some doubt about the correctness of the proof in the previous paper.

[1] In this paper, for the sake of ease of presentation and analysis, we assume that the factors are disjoint. However the ideas in this paper can easily be extended to non-disjoint factors. Indeed in the previous paper we defined a notion of conditional separability that can be applied to non-disjoint factors. This can naturally be extended to approximate conditional separability.

Separability is closely related to context-specific independence (CSI) [Boutilier *et al.*, 1996]. It can be viewed as CSI in which the context variable does not explicitly appear in the model. Separable models are also closely related to the Influence Model [Asavathiratham, 2000].

These ideas apply to the BK approach to monitoring in DBNs as follows:

**Definition 2.3:** The set of factors $\mathbf{X}_1, ..., \mathbf{X}_m$ is *self-sufficient* if $\cup_{i=1}^{m} \mathbf{X}_i = \mathbf{X}$, and for each $i$, $\mathbf{X}_1^-, ..., \mathbf{X}_m^-$ are sufficient for $\mathbf{X}_i$. ▮

Thus if the CPD for each factor is separable in terms of the factors at the previous time step, the factors are self-sufficient. Separability can be understood as characterizing the way information flows about a dynamic system. Intuitively, it says that information may flow from any of the factors at the previous time step to a factor at the current time step, but at any point in time only one of the factors is selected and information only flows from that one.

Given a self-sufficient set of factors, we can correctly propagate marginal probabilities from one time step to the next. Unfortunately, however, this is only true for propagating the marginals through the dynamics. Conditioning on observations breaks the property of sufficiency. The reason is that if we have factors $\mathbf{X}_1$ and $\mathbf{X}_2$ and have an observation about $\mathbf{X}_2$, we need to know what we can infer about $\mathbf{X}_1$. To know this, it is not enough to know the marginals over $\mathbf{X}_1$ and $\mathbf{X}_2$; we also need to know their joint distribution. In general we do not have the correct joint probability distribution between the factors using the BK approach, even for separable models. Thus propagating marginals with separable models is exact for the prediction task, where there are no observations about the future, but not for the monitoring task where we do have observations.

## 3 Approximate separability

These ideas lead naturally to two questions. The first question is, given the fact that separable models lead to correct propagation of marginals through the dynamics, do they also lead to more accurate monitoring? One may hope that because they are only subject to error when conditioning on observations, the total monitoring error will be lower. The second question addresses the problem of defining separable models in practice. While the previous paper gave some examples of separable models, in general real-world models will not be fully separable. Thus one may ask whether an approximate notion of separability leads to similar levels of performance to full separability.

**Definition 3.1:** $P(Z|XY)$ is *$\alpha$-separable* if there exist $0 \leq \alpha \leq 1$, $\gamma$, and CPDs $P_X$, $P_Y$ and $P_{XY}$, such that

$$\begin{aligned} P(Z|XY) &= \alpha(\gamma P_X(Z|X) + (1-\gamma)P_Y(Z|Y)) \\ &\quad +(1-\alpha)P_{XY}(Z|XY) \end{aligned}$$

▮

Trivially every CPD is 0-separable, and if it is $\alpha$-separable then it is also $\beta$-separable for $\beta < \alpha$.

**Definition 3.2 :** The *degree of separability* of $P(Z|XY)$ is the maximum $\alpha$ for which $P(Z|XY)$ is $\alpha$-separable. ▮

Similar to separability, we can understand approximate separability intuitively as characterizing the flow of information about the system. If $P(Z|XY)$ is highly separable, it says that while information can flow from both $X$ and $Y$ to $Z$, most of the time it only flows from one of them or the other. While it may just as often flow from $X$ or from $Y$ (e.g. if $\gamma = 1/2$), on only rare occasions does information flow from both of them together. Approximate separability can also be viewed as an approximate form of CSI.

**Example 3.3:** The following is an example of an approximately separable decomposition:

| | | $X$ | |
|---|---|---|---|
| $X^-$ | $Y^-$ | F | T |
| F | F | .9 | .1 |
| F | T | .99 | .01 |
| T | F | .1 | .9 |
| T | T | .01 | .99 |

=

.91

| | $X$ | |
|---|---|---|
| $X^-$ | F | T |
| F | .989 | .011 |
| T | .011 | .989 |

\+ .09

| | | $X$ | |
|---|---|---|---|
| $X^-$ | $Y^-$ | F | T |
| F | F | 0 | 1 |
| F | T | 1 | 0 |
| T | F | 1 | 0 |
| T | T | 0 | 1 |

In this example, $X$ tends to take on the same value as $X^-$ with high probability. However, $Y^-$ also has an influence, and the influence of $Y^-$ is flipped depending on $X^-$. If $X^-$ is F, $Y^-$ being T makes $X$ more likely to be F, but if $X^-$ is T, $Y^-$ being T makes $X$ more likely to be T. The fact that in the decomposition there is no $P_Y$ is a coincidence. However the fact that $P_{XY}$ is a XOR model is not. ▮

In fact, approximate separability is quite a natural property that holds for many models.

**Definition 3.4 :** $P(X|X^-Y^-)$ is $\kappa$-persistent if $P(x|x^-y^-) \geq \kappa$ whenever $x = x^-$, for all values $y^-$ of $Y^-$. ▮

Persistence is the property that things tend to stay the same as before. This is a natural property of dynamic systems. It does not say that a different factor has only a small influence on a given factor in relative terms — $X$ may be many times more likely to change given some values of $Y^-$ than others.

**Proposition 3.5:** *If $P(X|X^-Y^-)$ is $\kappa$-persistent, it is $\kappa$-separable.*

**Proof:** $P(X|X^-Y^-)$ can be decomposed into $\kappa I(X|X^-) + (1-\kappa)P_{XY}(X|X^-Y^-)$ where $I$ is the identity CPD, and

$$P_{XY}(x|x^-y^-) = \begin{cases} \frac{P(x|x^-y^-)}{1-\kappa} & \text{if } x \neq x^- \\ \frac{P(x|x^-y^-)-\kappa}{1-\kappa} & \text{if } x = x^- \end{cases}$$

By the definition of $\kappa$-persistence,
$0 \leq P_{XY}(x|x^-y^-) \leq 1$, and
$\sum_x P_{XY}(x|x^-y^-) = \frac{1-\kappa}{1-\kappa} = 1$. ▮

Note that the converse does not hold: $P(X|X^-Y^-)$ may be much more than $\kappa$-separable, and there are many ways to get approximately separable models other than through persistence.

## 4 Experiments

Figure 1 (a) shows the performance of various models on the prediction task with no observations. The $x$-axis shows the degree of non-separability of the model: the left endpoint is a fully separable model, while the right endpoint is a completely non-separable model. The experiments are performed on a small model with two factors $X$ and $Y$ and a variable $Z$ which is a noisy observation of $Y$. The figure shows the KL-distance between the true marginal over $X$ and the approximate marginal obtained using the BK approach, averaged over 10000 runs with random parameters. As the theory predicts, fully separable models have zero error. What is interesting, though, is the shape of the curve. We see that approximately separable models also have very low error; even 50%-separability leads to good performance.

Figure 1 (b) shows the performance of the different models on the monitoring task. Again the results are averaged over 10000 runs. Here the news is very good. Although the theory does not predict that separable models will perform well, in fact they have very small error. The curve has a similar shape; again approximately separable models have small error.

One natural hypothesis as to why separable models perform so well in these examples is that the factors are less dependent on each other. This turns out not to be the case. Figure 1 (c) shows the degree of dependence between the factors, measured by the KL distance between the true joint distribution and the product of its marginals. This is computed over the same kind of models as for Figure 1 (a) and Figure 1 (b). The lowest degree of dependence is actually attained for models with an intermediate degree of separability.

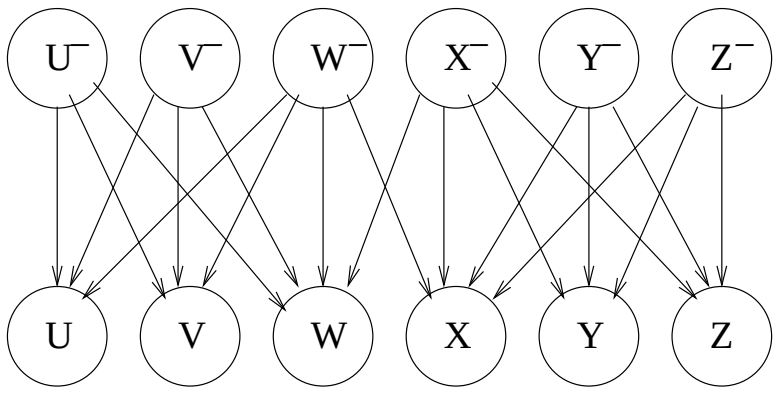

Figure 2: A six node DBN

**Example 4.1:** This example shows how using separability can lead to much better factorization of a dynamic system. Figure 2 shows the structure of a six-node DBN. The variables $U$,$V$ and $W$ are all dependent on $U^-$,$V^-$ and $W^-$; similarly $X$,$Y$ and $Z$ are all dependent on $X^-$,$Y^-$ and $Z^-$. The only other interactions are between $W^-$ and $X$ and between $X^-$ and $W$. This structure suggests a natural factorization into two factors: $UVW$ and $XYZ$. However, suppose the CPD of $X$ is as follows:

| | | | | X | |
|---|---|---|---|---|---|
| $X^-$ | $Y^-$ | $Z^-$ | $W^-$ | F | T |
| F | F | F | F | 0.9 | 0.1 |
| F | F | F | T | 0.5 | 0.5 |
| F | F | T | F | 0.7 | 0.3 |
| F | F | T | T | 0.3 | 0.7 |
| F | T | F | F | 0.7 | 0.3 |
| F | T | F | T | 0.3 | 0.7 |
| F | T | T | F | 0.5 | 0.5 |
| F | T | T | T | 0.1 | 0.9 |
| T | F | F | F | 0.5 | 0.5 |
| T | F | F | T | 0.9 | 0.1 |
| T | F | T | F | 0.3 | 0.7 |
| T | F | T | T | 0.7 | 0.3 |
| T | T | F | F | 0.3 | 0.7 |
| T | T | F | T | 0.7 | 0.3 |
| T | T | T | F | 0.1 | 0.9 |
| T | T | T | T | 0.5 | 0.5 |

This CPD is highly separable in terms of $X^-W^-$ and $Y^-Z^-$, but highly non-separable in terms of $X^-Y^-Z^-$ and $W^-$. (The reason will be explained in Section 5. Note that this example is not highly persistent.) If the other CPDs are similarly designed, it turns out that $\{UV, WX, YZ\}$ are a self-sufficient set of factors, while $\{UVW, XYZ\}$ are not. This suggests that $\{UV, WX, YZ\}$ may be a better factorization. Note that it is not the case that $Y^-$ and $Z^-$ together have smaller influence on $X$ than $W^-$. There-fore there is no reason to prefer this factorization based on considerations of degree of influence.

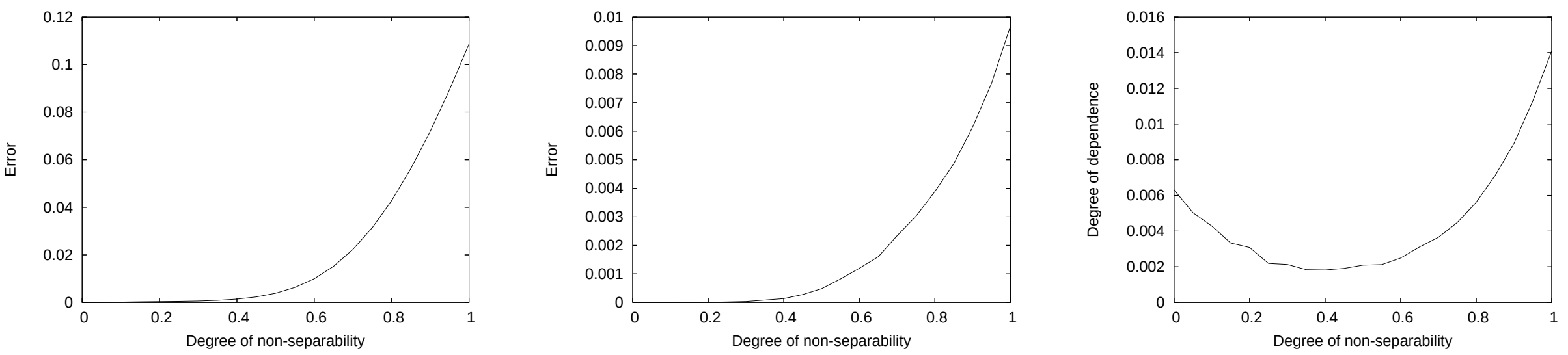

(a) (b) (c)

Figure 1: Error as a function of degree of non-separability: (a) without observations; (b) with observations; (c) degree of dependence between the factors.

The monitoring results, with $Z$ being observed, are as follows: the average absolute error in $P(U = T)$ for the $\{UVW, XYZ\}$ factorization is 0.038, while for the $\{UV, WX, YZ\}$ factorization it is 0.018. Similarly the average KL distance in the marginal of $U$ is 0.007 for the $\{UVW, XYZ\}$ factorization and 0.002 for the $\{UV, WX, YZ\}$ factorization. Thus a factorization based on separability can lead to much better performance than the "obvious" one. ▮

## 5 Stucture of approximately separable decompositions

Two natural questions arise immediately: what do approximately separable models look like, and why do they perform well, at least on the examples studied? In this section we analyze the structure of approximately separable decompositions, by presenting a method to determine the degree of separability of a model. The notation $z_i$ refers to the $i$-th possible value of $Z$. The goal is, given a CPD $P(Z|XY)$, to find an optimal approximately separable decomposition. In general, this can be formulated as a linear programming problem:

$$
\begin{array}{lcl}
\max & \alpha_1 + \alpha_2 & \\
\text{s.t.} & \alpha_1 p_{ij} + \alpha_2 q_{ik} + \alpha_3 r_{ijk} = P(z_i|x_j y_k) & \forall i,j,k \\
& \alpha_1 + \alpha_2 + \alpha_3 = 1 & \\
& \alpha_3 \geq 0 & \\
& 0 \leq p_{ij} \leq 1 & \forall i,j \\
& 0 \leq q_{ik} \leq 1 & \forall i,k \\
& 0 \leq r_{ijk} \leq 1 & \forall i,j,k \\
& \sum_i p_{ij} = 1 & \forall j \\
& \sum_i q_{ik} = 1 & \forall k \\
& \sum_i r_{ijk} = 1 & \forall j,k
\end{array}
$$

From the solution, we can identify $\alpha = \alpha_1 + \alpha_2$, $\gamma = \frac{\alpha_1}{\alpha}$, $P_X(z_i|x_j) = p_{ij}$, $P_Y(z_i|y_k) = q_{ik}$, and $P_{XY}(z_i|x_j y_k) = r_{ijk}$. Note that there is no constraint that $\alpha_1$ or $\alpha_2$ be greater than 0. This is because $\gamma$ is not constrained to be between 0 and 1. However we do have a constraint that $\alpha_3 \geq 0$. Since $\alpha$ is the degree of separability, it does not make sense to say that a model is more than completely separable. In special cases, the program can be solved directly. Examining these special cases yields insights into the structure of approximately separable decompositions.

**Case 1:** $X$,$Y$ and $Z$ are binary.

We can write down the equations defining $\alpha$-separability as

$$P(z_1|x_1, y_1) = \alpha\gamma P_X(z_1|x_1) + \alpha(1-\gamma)P_Y(z_1|y_1) + (1-\alpha)P_{XY}(z_1|x_1 y_1) \tag{1}$$

$$P(z_1|x_1, y_2) = \alpha\gamma P_X(z_1|x_1) + \alpha(1-\gamma)P_Y(z_1|y_2) + (1-\alpha)P_{XY}(z_1|x_1 y_2) \tag{2}$$

$$P(z_1|x_2, y_1) = \alpha\gamma P_X(z_1|x_2) + \alpha(1-\gamma)P_Y(z_1|y_1) + (1-\alpha)P_{XY}(z_1|x_2 y_1) \tag{3}$$

$$P(z_1|x_2, y_2) = \alpha\gamma P_X(z_1|x_2) + \alpha(1-\gamma)P_Y(z_1|y_2) + (1-\alpha)P_{XY}(z_1|x_2 y_2) \tag{4}$$

Since $P(z_2|x_j y_k) = 1 - P(z_1|x_j y_k)$, the corresponding equations for $z_2$ automatically hold when these hold. Subtracting (1) from (2) and (3) from (4), we get

$$
\begin{aligned}
&P(z_1|x_1 y_2) - P(z_1|x_1 y_1) \\
&\quad = \alpha(1-\gamma)(P_Y(z_1|y_2) - P_Y(z_1|y_1)) \\
&\qquad + (1-\alpha)(P_{XY}(z_1|x_1 y_2) - P_{XY}(z_1|x_1 y_1))
\end{aligned} \tag{5}
$$

$$
\begin{aligned}
&P(z_1|x_2 y_2) - P(z_1|x_2 y_1) \\
&\quad = \alpha(1-\gamma)(P_Y(z_1|y_2) - P_Y(z_1|y_1)) \\
&\qquad + (1-\alpha)(P_{XY}(z_1|x_2 y_2) - P_{XY}(z_1|x_2 y_1))
\end{aligned} \tag{6}
$$

We see immediately that for a fully separable model, it must hold that

$$P(z_1|x_1 y_2) - P(z_1|x_1 y_1) = P(z_1|x_2 y_2) - P(z_1|x_2 y_1) \tag{7}$$

In other words, $Y$ must have exactly the same effect on $Z$, no matter what the state of $X$. If we examine the CPD in Figure 4.1, we see that this holds for the decomposition into $\{X^- W^-\}$ and $\{Y^- Z^-\}$, but not for the decomposition into $\{W^-\}$ and $\{X^- Y^- Z^-\}$. If

(7) does not hold, we must have $\alpha < 1$. In this case, subtracting (5) from (6), we get

$$\begin{aligned} & P(z_1|x_2y_2) - P(z_1|x_2y_1) \\ & -P(z_1|x_1y_2) + P(z_1|x_1y_1) \\ & \quad = \; (1-\alpha)(P_{XY}(z_1|x_2y_2) - P_{XY}(z_1|x_2y_1) \\ & \qquad\qquad - P_{XY}(z_1|x_1y_2) + P_{XY}(z_1|x_1y_1)) \end{aligned} \tag{8}$$

If the left hand side of (8) is positive, $\alpha$ is maximized when we set

$$\begin{aligned} P_{XY}(z_1|x_2y_2) = P_{XY}(z_1|x_1y_1) = 1 \\ P_{XY}(z_1|x_2y_1) = P_{XY}(z_2|x_1y_2) = 0 \\ \alpha = 1 - \tfrac{1}{2}(P(z_1|x_2y_2) - P(z_1|x_2y_1) \\ - P(z_1|x_1y_2) + P(z_1|x_1y_1)) \end{aligned}$$

We get a symmetric result when the left hand side of (8) is negative. We see that $P_{XY}$ must either be an equality test or a XOR of $X$ and $Y$. Thus the pattern in Example 3.3 is no coincidence. Once we have the value of $\alpha$, we can recover $\gamma$, $P_Y$ and $P_X$. It can be shown that there always exists a feasible solution for these variables, though $\gamma$ might have to be negative.

**Case 2:** $X$ and $Y$ are binary, while $Z$ has $n$ values.

We can write down equations $(1_i)$ to $(4_i)$ similar to (1) to (4) above for all values $z_i$ of $Z$. We then subtract $(1_i)$ from $(2_i)$ and $(3_i)$ from $(4_i)$ to get $(5_i)$ and $(6_i)$. We then subtract $(5_i)$ from $(6_i)$ to get $(8_i)$, as above:

$$\begin{aligned} & P(z_i|x_2y_2) - P(z_i|x_2y_1) \\ & -P(z_i|x_1y_2) + P(z_i|x_1y_1) \\ & \quad = \; (1-\alpha)(P_{XY}(z_i|x_2y_2) - P_{XY}(z_i|x_2y_1) \\ & \qquad\qquad - P_{XY}(z_i|x_1y_2) + P_{XY}(z_i|x_1y_1)) \end{aligned}$$

Let $A_i$ be the left hand side of $(8_i)$. Now, the expression

$$\begin{aligned} P_{XY}(z_i|x_2y_2) - P_{XY}(z_i|x_2y_1) - \\ P_{XY}(z_i|x_1y_2) + P_{XY}(z_i|x_1y_1) \end{aligned}$$

must be proportional to $A_i$. Furthermore, if we consider only positive $A_i$, in order to maximize $\alpha$ we must maximize this expression. To do this, we first set

$$P_{XY}(z_i|x_2y_1) = P_{XY}(z_i|x_1y_2) = 0$$

We must then maximize $P_{XY}(z_i|x_2y_2)$ and $P_{XY}(z_i|x_1y_1)$ subject to the constraints that their sum is proportional to $A_i$, and $\sum_i P_{XY}(z_i|x_2y_2) \leq 1$ and $\sum_i P_{XY}(z_i|x_1y_1) \leq 1$. This is achieved by setting

$$\begin{aligned} & G = \textstyle\sum_{i:A_i>0} A_i \\ & P_{XY}(z_i|x_2y_2) = P_{XY}(z_i|x_1y_1) = \tfrac{A_i}{G} \\ & \alpha = 1 - \tfrac{G}{2} \end{aligned}$$

Now let us consider negative $A_i$. Since $\sum_i P(z_i|x_jy_k)$ must be 1, it follows that the sum of the positive $A_i$ equals the absolute sum of the negative $A_i$. Therefore we can use the same $G$, and set, symmetrically

$$\begin{aligned} P_{XY}(z_i|x_2y_2) = P_{XY}(z_i|x_1y_1) = 0 \\ P_{XY}(z_i|x_1y_2) = P_{XY}(z_i|x_2y_1) = -\frac{A_i}{G} \end{aligned}$$

Intuitively, $A_i$ is the degree of deviation from the situation in which the influence of $X$ does not depend on the state of $Y$ (and vice versa). $\alpha$ is determined by $G$, which is the sum of all the positive deviations. Therefore the deviations for different states of $Z$ sum together to limit the degree of separability. It would seem, at first sight, that if $Z$ has more states, there will be more deviations to sum together, and it will be more difficult to attain a high degree of separability. However, when $Z$ has many states, the probability mass for $P(Z|XY)$ will be divided up among more states, so we would expect the individual deviations to be smaller. Thus in balance, it should be equally difficult to achieve a high degree of separability whether $Z$ has few or many states.

**Case 3:** $X$ and $Z$ are binary, while $Y$ has $n$ values.

We write down the following equations for each $y_k$:

$$\begin{aligned} P(z_1|x_1y_k) &= \alpha(\gamma P_X(z_1|x_1) + (1-\gamma)P_Y(z_1|y_k)) \\ &\quad + (1-\alpha)P_{XY}(z_1|x_1y_k) \\ P(z_1|x_2y_k) &= \alpha(\gamma P_X(z_1|x_2) + (1-\gamma)P_Y(z_1|y_k)) \\ &\quad + (1-\alpha)P_{XY}(z_1|x_2y_k) \end{aligned}$$

By subtracting, we get for $k = 1, ..., n-1$

$$\begin{aligned} & P(z_1|x_1y_{k+1}) - P(z_1|x_1y_k) \\ & \quad = \; \alpha(1-\gamma)(P_Y(z_1|y_{k+1}) - P_Y(z_1|y_k)) \\ & \qquad + (1-\alpha)(P_{XY}(z_1|x_1y_{k+1}) - P_{XY}(z_1|x_1y_k)) \\ & P(z_1|x_2y_{k+1}) - P(z_1|x_2y_k) \\ & \quad = \; \alpha(1-\gamma)(P_Y(z_1|y_{k+1}) - P_Y(z_1|y_k)) \\ & \qquad + (1-\alpha)(P_{XY}(z_1|x_2y_{k+1}) - P_{XY}(z_1|x_2y_k)) \end{aligned}$$

Subtracting again, we get

$$\begin{aligned} & P(z_1|x_2y_{k+1}) - P(z_1|x_2y_k) \\ & -P(z_1|x_1y_{k+1}) + P(z_1|x_1y_k) \\ & \quad = \; (1-\alpha)(P_{XY}(z_1|x_2y_{k+1}) - P_{XY}(z_1|x_2y_k) \\ & \qquad\qquad - P_{XY}(z_1|x_1y_{k+1}) + P_{XY}(z_1|x_1y_k)) \end{aligned}$$

For $k = 1, ..., n-1$, let

$$\begin{aligned} A_k = P(z_1|x_2y_{k+1}) - P(z_1|x_2y_k) \\ - P(z_1|x_1y_{k+1}) + P(z_1|x_1y_k) \end{aligned} \tag{9}$$

Intuitively, $A_k$ is the deviation from the situation in which the influence of $X$ does not depend on whether $Y$ is equal to $y_k$ or $y_{k+1}$. We get to choose, for $k = 1, ..., n$

$$B_k = P_{XY}(z_1|x_2y_k) - P_{XY}(z_1|x_1y_k)$$

Now, if we choose $B_1$, that determines all the other $B_k$, because

$$A_k \quad = \quad (1-\alpha)(B_{k+1} - B_k)$$

from which

$$B_{k+1} = B_k - \frac{A_k}{1-\alpha}. \tag{10}$$

We can now derive constraints on the $B_k$, which will lead in particular to constraints on $B_1$. Using the fact that $B_k$ must be between $-1$ and 1, we first get for $B_n$

$$-1 \leq B_n \leq 1$$

Then, for $B_{n-1}$, we combine the constraint on $B_n$ using (10), and add a new direct constraint:

$$\begin{array}{ccccc} -1 & \leq & B_{n-1} & \leq & 1 \\ -1 & \leq & B_{n-1} - \frac{A_{n-1}}{1-\alpha} & \leq & 1 \end{array}$$

Then, working our way down to $B_1$, we add one new constraint at each step, finally getting

$$\begin{array}{ccccc} -1 & \leq & B_1 & \leq & 1 \\ -1 & \leq & B_1 - \frac{A_1}{1-\alpha} & \leq & 1 \\ & & \cdots & & \\ -1 & \leq & B_1 - \frac{\sum_{k=1}^{n-1} A_k}{1-\alpha} & \leq & 1 \end{array} \tag{11}$$

It is clear that if all the constraints on $B_1$ are satisfied, then so are the constraints on higher $B_k$, so we only have to satisfy those. Now, let $C_k = A_1 + \ldots + A_k$, i.e. $C_k$ is a partial sum of deviations. Each $C_k$, from $k = 0$ to $n-1$ appears in one of the constraints (11). The maximum positive $C_k$ and the minimum negative $C_k$ place the strongest constraints on $B_1$. Let $C^* = \max(0, \max_{k=1}^{n-1} C_k)$, and $C_* = -\min(0, \min_{k=1}^{n-1} C_k)$. If we ensure that

$$\begin{array}{l} B_1 - \frac{C^*}{1-\alpha} \geq -1 \\ B_1 + \frac{C_*}{1-\alpha} \leq 1 \end{array} \tag{12}$$

all the constraints will be satisfied. $\alpha$ is maximized when constraints (12) are tight. This occurs when $\alpha = 1 - \frac{C^*+C_*}{2}$ and $B_1 = \frac{C^*-C_*}{C^*+C_*}$.

To see why this result makes sense, note that from (9),

$$\begin{array}{l} C_k = P(z_1|x_2 y_{k+1}) - P(z_1|x_2 y_1) \\ \qquad - P(z_1|x_1 y_{k+1}) + P(z_1|x_1 y_1) \end{array}$$

This is the degree to which the influence of $X$ changes when $Y$ is equal to $y_{k+1}$ compared to $y_1$. It is the maximum and minimum of these changes that determines $\alpha$. The interesting, and encouraging, thing is that it is not the sum of the positive, and absolute sum of the negative, deviations that determine $\alpha$. We might have thought that if there were several positive $A_k$, their influence would accumulate and $\alpha$ would necessarily be reduced. If that were the case, the degree of separability would be inversely proportional to the number of values of $Y$. Instead, $\alpha$ depends on the maximum and minimum deviations. We would still expect $\alpha$ to decrease with the number of values of $Y$, but not to the same degree.

# 6 Why separable models perform well

## 6.1 Error bound

To attempt to answer the question of why separable models appear to perform so well, we first prove a bound on the expected error for a very special case of separable models. These are models with two hidden binary variables $X$ and $Y$, that both depend on $X^-$ and $Y^-$, and a binary observation $Z$ that depends only on $Y$. $P(X \mid X^- Y^-)$ and $P(Y \mid X^- Y^-)$ are separable, with parameterization:

$$P(X \mid X^- Y^-) = \gamma_X P_X^X(X \mid X^-) + (1-\gamma_X) P_X^Y(X \mid Y^-)$$

$$P(Y \mid X^- Y^-) = \gamma_Y P_Y^X(Y \mid X^-) + (1-\gamma_Y) P_Y^Y(Y \mid Y^-)$$

$$P(Z \mid Y) = P_Z(Z \mid Y)$$

The error bound depends on a number of quantities characterizing the system. We define the following quantities that characterize the degree to which the value of the parent $X$ or $Y$ at the previous time step influences the child at the current time step:

$$\begin{array}{lcl} \lambda_X^X & = & |P_X^X(x_2 \mid x_2) - P_X^X(x_2 \mid x_1)| \\ \lambda_X^Y & = & |P_X^Y(x_2 \mid y_2) - P_X^Y(x_2 \mid y_1)| \\ \lambda_Y^X & = & |P_Y^X(y_2 \mid x_2) - P_Y^X(y_2 \mid x_1)| \\ \lambda_Y^Y & = & |P_Y^Y(y_2 \mid y_2) - P_Y^Y(y_2 \mid y_1)| \end{array}$$

We also define the following quantity that characterizes the degree to which the evidence is informative:

$$\lambda_Z \quad = \quad |P_Z(z_2 \mid y_2) - P(z_2 \mid y_1)|$$

The following quantities also turn out to be useful:

$$\begin{array}{lcl} \lambda_{XY}^X & = & \left| \begin{array}{l} P_X^X(x_2 \mid x_2) P_Y^X(y_2 \mid x_2) \\ -P_X^X(x_2 \mid x_1) P_Y^X(y_2 \mid x_1) \end{array} \right| \\ \lambda_{XY}^Y & = & \left| \begin{array}{l} P_X^Y(x_2 \mid x_2) P_Y^Y(y_2 \mid x_2) \\ -P_X^Y(x_2 \mid x_1) P_Y^Y(y_2 \mid x_1) \end{array} \right| \\ \zeta^X & = & \max(\lambda_X^X, \lambda_Z(\lambda_X^X - 2\lambda_{XY}^X + 2\lambda_Y^X)) \\ \zeta^Y & = & \max(\lambda_X^Y, \lambda_Z(\lambda_X^Y - 2\lambda_{XY}^Y + 2\lambda_Y^Y)) \end{array}$$

Let $\mu$ denote $P(X = T, Y = T)$ under the true joint distribution at a particular time point. Let $\mu_X = P(X = T)$ and $\mu_Y = P(Y = T)$ under the true marginals at the time point, and let $\hat{\mu}_X = \hat{P}(X = T)$ and $\hat{\mu}_Y = \hat{P}(Y = T)$ under the approximate marginals maintained by the BK algorithm. We define two error measures. The first, called the *joint error*, denoted $\delta$, is the $\ell^\infty$ distance (i.e. max norm) between the true joint distribution at the current time point and the product of its marginals. The second, called the *marginal error*, denoted $\delta_X$ (respectively $\delta_Y$), is the $\ell^\infty$ distance between the true marginal over $X$ (resp. $Y$) at the time point and the approximate marginal produced by the BK method.

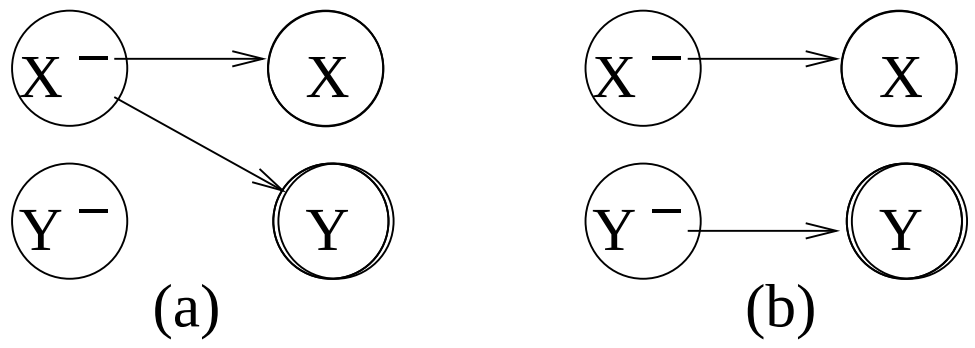

Figure 3: Types of move for simple separable model

In the following theorem, the bounding expressions are complicated and not all that interesting. However they do provide quite good bounds on the marginal error. Also, the intuition behind the proof is informative, and the key lemmas describe how the error develops.

**Theorem 6.1:** *For a system as defined above, at all time points, taking the expectation over all observation sequences up to the time point,*

$$\begin{array}{rcl} E[\delta] & \leq & H \\ E[\delta_X] & \leq & J \\ E[\delta_Y] & \leq & K \end{array}$$

*where*

$$\begin{array}{rcl}
H & = & \frac{(1-\lambda_Z^2)L}{4(1-(1-\lambda_Z^2)M)} \\
J & = & \frac{2H(1-(1-\gamma_Y)\lambda_Y^Y)\lambda_Z M}{N} \\
K & = & \frac{2H\gamma_Y\lambda_Y^X\lambda_Z M}{N} \\
L & = & \gamma_X\gamma_Y\lambda_X^X\lambda_Y^X + (1-\gamma_X)(1-\gamma_Y)\lambda_X^Y\lambda_Y^Y \\
M & = & \gamma_X(1-\gamma_Y)\lambda_X^X\lambda_Y^Y + (1-\gamma_X)\gamma_Y\lambda_X^Y\lambda_Y^X \\
N & = & (1-(1-\gamma_Y)\lambda_Y^Y)(1-\gamma_X O) \\
 & & -(1-\gamma_X)\gamma_Y\lambda_Y^X P \\
O & = & \gamma_Y\zeta^X + (1-\gamma_Y)\lambda_X^X \\
P & = & \gamma_Y\lambda_X^Y + (1-\gamma_Y)\zeta^Y
\end{array}$$

*provided the errors at the initial time point are at most these quantities.*

Based on this theorem, we can produce an average bound on the error in the marginal over $X$, taken over 10000 random parameterizations, of 0.000662. This compares with a true average error of 0.000240. Thus the bound is less than three times the actual error.

The proof of this result is based on the idea that in a separable model, the system makes two kinds of moves, illustrated in Figure 3. In the first kind of move, both variables at the current time point depend on the same variable at the previous time point. In the second kind of move, they both depend on different variables at the previous time point. In addition to the moves shown in the figure, there are two more symmetric moves. While any of the four moves may be chosen at any point in time, only one of the moves will actually be chosen, and the process switches between the different kinds of moves. The key point is that in the first kind of move, all previous joint error is forgotten, because only one variable at the previous time point affects the current state. In the second type of move, new marginal error may be introduced by ignoring dependencies between the variables when conditioning on the observation. However, no new joint error will be introduced. These two effects combine to keep the marginal error small. In particular, in each kind of move the error is characterized as follows:

**Lemma 6.2:** *For the first kind of move, taking the expectation over observations,* $E[\delta] \leq \frac{1}{4}\lambda_X^X\lambda_Y^X(1-\lambda_Z^2)$.

Thus the joint error after the first kind of move does not depend on the previous error, and the new error depends on the degree to which the old parent $X$ influences both $X$ and $Y$. This makes sense. The more both $X$ and $Y$ depend on the previous $X$, the more dependence will be introduced between them. The new error is also less the more informative the observation. This is somewhat surprising, but makes sense when we consider that perfectly informative evidence serves to completely decouple $X$ from $Y$.

**Lemma 6.3:** *For the first kind of move,* $E[\delta_X] \leq \zeta^X\delta_X^-$.

**Lemma 6.4:** *For the first kind of move,* $E[\delta_Y] \leq \lambda_Y^X\delta_X^-$.

Thus the new marginal error of both $X$ and $Y$ depends on the old marginal error of $X$. The weaker the influence of the old $X$, the more the old error is forgotten. As it turns out, the marginal error of $X$ depends on a more complicated expression, but $\lambda_X^X$ is still one of its main components.

**Lemma 6.5:** *For the second kind of move,* $E[\delta] \leq \lambda_X^X\lambda_Y^Y(1-\lambda_Z^2)\delta^-$.

Thus the new joint error depends on the old joint error. The less the influence of the old parents on their respective children, and the more informative the observation, the quicker the old error is forgotten.

**Lemma 6.6:** *For the second kind of move,* $E[\delta_X] \leq 2\lambda_X^X\lambda_Y^Y\lambda_Z\delta^- + \lambda_X^X\delta_X^-$.

The new marginal error of $X$ is made up of two components. One component results from the previous joint error, i.e. it results from ignoring the previous dependencies between the factors. This component decreases with the lack of influence of the old parents on their respective children, and the uninformativeness of the observation. Note that the effect of the observation is different from before. Now uninformative observations lead to lower error. The second component of the marginal error is a result of the previous marginal error, which is forgotten in proportion to the lack of influence of the old $X$ on the new $X$.

**Lemma 6.7:** *For the second kind of move,* $E[\delta_Y] \leq \lambda_Y^Y \delta_Y^-$.

### 6.2 Sources of Error

A second analysis examines the different sources of error that arise for the BK approach. We identify two sources of error. In the following, we present the two sources of error for a system with two factors, but the concepts generalize to more than the two factors. Our experiments only consider the two-factor case.

The first source of error, called Type A, results from not taking into account previous dependencies between the factors when propagating through the dynamics to obtain the prior distribution at the current time point. Let $\hat{\mu}^-$ be the approximate posterior distribution over the previous state, and $\hat{\mu}_X^-$ and $\hat{\mu}_Y^-$ be its marginals. Type A error is the error that results from instead of computing the prior

$$\varphi'(XY) = \sum_{x^- y^-} P(XY | x^- y^-) \hat{\mu}^-(x^- y^-),$$

approximating it with

$$\tilde{\varphi}(XY) = \sum_{x^- y^-} P(XY | x^- y^-) \hat{\mu}_X^-(x^-) \hat{\mu}_Y^-(y^-)$$

The Type A error is accumulated by repeatedly applying this approximation at every iteration, but assuming the correct dependencies are used when conditioning on observations.

The second source of error, called Type B, results from not taking into account previous dependencies between the factors when conditioning on the observations to obtain the posterior distribution. It must be pointed out that some dependencies between the factors are taken into account when conditioning. These are dependencies introduced by a single step of the dynamics, due to the fact that different variables depend on the same variable at the previous time step. It is only old dependencies that held between the factors at the previous time point that are not taken into account. Type B error is produced when the prior is computed using the correct degree of dependence, but when conditioning on the observations only the dependence introduced in $\tilde{\varphi}$ is used.

Note that separable models only have Type B error. They have no Type A error because the degree of dependence at the previous time point does not affect the prior marginals at the current time point.

To analyze the sources of error, we define a process for each source that only has that source of error. The following procedure isolates Type A error:

1. Determine degree of dependence $d$:
   (a) Propagate $\mu^-$, the true posterior at the previous time point, through the dynamics, to obtain the true prior $\varphi$.
   (b) Let $\varphi_X$ and $\varphi_Y$ be the marginals of $\varphi$.
   (c) Let $d = \varphi - \varphi_X \varphi_Y$. Thus $d$ measures the dependence between the factors in the true prior at the current time point.
2. Propagation process:
   (a) Propagate $\hat{\mu}_X^- \hat{\mu}_Y^-$ to obtain $\tilde{\varphi}$, as described above.
   (b) Let $\tilde{\varphi}_X$ and $\tilde{\varphi}_Y$ be the marginals of $\tilde{\varphi}$. Thus these are the prior marginals obtained by beginning with the approximate posterior at the previous time point, and assuming the factors were independent. These marginals incorporate Type A error.
   (c) Let $\varphi^* = \tilde{\varphi}_X \tilde{\varphi}_Y + d$. Thus $\varphi^*$ uses the incorrect prior marginals, but has the correct degree of dependence between them.
   (d) Condition $\varphi^*$ on the observation to obtain the approximate posterior $\hat{\mu}$. Thus the correct dependence between the factors is used when conditioning on the observation.
   (e) Project $\hat{\mu}$ onto $\hat{\mu}_X$ and $\hat{\mu}_Y$ and measure the KL distance between the true posterior marginal $\mu_X$ and $\hat{\mu}_X$.
3. Repeat for the next iteration.

The following is the process for Type B error:

1. Compute degree of dependences $d^-$ and $\tilde{d}$:
   (a) Let $d^- = \mu^- - \mu_X^- \mu_Y^-$. Thus $d^-$ measures old dependencies between the factors according to the true posterior at the previous time point.
   (b) Let $\tilde{\varphi}$ be the prior distribution resulting from propagating $\hat{\mu}_X^- \hat{\mu}_Y^-$ through the dynamics, as before.

(c) Let $\tilde{d} = \tilde{\varphi} - \tilde{\varphi}_X \tilde{\varphi}_Y$. Thus $\tilde{d}$ measures the dependence introduced by a single step of the dynamics.

2. Propagation process:

   (a) Let $\mu^{*-} = \hat{\mu}_X^- \hat{\mu}_Y^- + d^-$. This uses the approximate posterior marginals at the previous time point, but incorporates the correct degree of dependence. This degree of dependence is taken into account when propagating through the dynamics.

   (b) Propagate $\hat{\mu}^-$ through the dynamics to obtain an approximate prior $\varphi'$. (This is slightly different from the $\varphi'$ defined earlier, because it incorporates the true degree of dependence contained in $\mu^-$, rather than the approximate degree of dependence contained in $\hat{\mu}^-$.)

   (c) Project $\varphi'$ to obtain approximate prior marginals $\varphi'_X$ and $\varphi'_Y$.

   (d) Let $\hat{\varphi} = \varphi'_X \varphi'_Y + \tilde{d}$. Only single-step dependencies are taken into account for conditioning, thus producing Type B error.

   (e) Condition $\hat{\varphi}$ on the observation to obtain $\hat{\mu}$.

   (f) Project $\hat{\mu}$ to obtain $\hat{\mu}_X$ and $\hat{\mu}_Y$, and measure the KL distance between $\mu_X$ and $\hat{\mu}_X$.

3. Repeat for the next iteration.

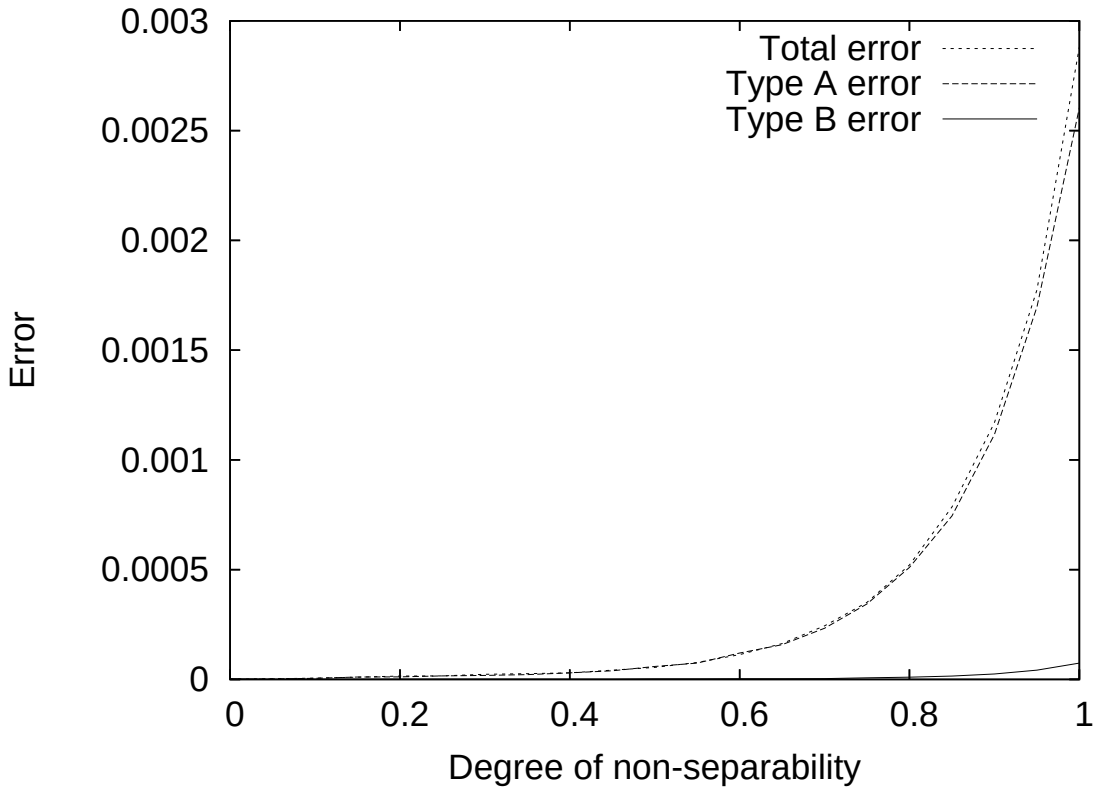

Figure 4: Contributions of sources of error.

The two types of error are measured using these processes, and the results are shown in Figure 4. We see that Type A error is almost equal to the total error, while Type B error is a small fraction of the total error, even for the most non-separable models. It seems that taking into account old dependencies when conditioning on observations is not that important. Since separable models only have this source of error, this goes some way towards explaining why they do well.

# 7 Conclusion

This paper has presented approximate separability as a characterization of weak interaction in dynamic systems. Approximate separability leads to accurate propagation of marginals using the factored approach to monitoring. We have analyzed the structure of approximately separable models and provided some explanation as to why they perform well.

More experimentation is needed on a wider array of examples to strengthen the conclusions in this paper. In addition, it would be interesting to see if other forms of weak interaction, such as noisy-or, also allow BK to work well. An important open issue is defining a notion of approximate separability for continuous variables. Another issue to explore is fully exploiting approximate separability to improve efficiency, perhaps extending the ideas of [Poole and Zhang, 2003]. It would also be useful to find a way to decompose a system automatically into weakly interacting factors.